\documentclass[runningheads]{llncs}
\usepackage[T1]{fontenc}

\usepackage{amsmath} 
\usepackage{graphicx,verbatim}

\usepackage{diagbox}
\usepackage{multirow}

\usepackage{tabularx,booktabs}
\newcolumntype{P}[1]{>{\centering\arraybackslash}p{#1}}
\usepackage[
colorlinks=true,
urlcolor=blue,
linkcolor=black,
citecolor=black
]{hyperref}
\newcolumntype{Y}{>{\centering\arraybackslash}X}

\newcommand{\repeatthanks}{\textsuperscript{\thefootnote}}

\begin{document}
\title{Neural Cellular Automata for Weakly Supervised Segmentation of White Blood Cells}
\titlerunning{Neural Cellular Automata for Weakly Supervised Segmentation}
%
\author{Michael Deutges\inst{1} \and
Chen Yang \inst{2} \and
Raheleh Salehi \inst{1,3} \and
Nassir Navab \inst{4} \and \\
Carsten Marr \inst{1}\thanks{Co-corresponding: \{carsten.marr,ario.sadafi\}@helmholtz-munich.de} \and
Ario Sadafi \inst{1,4}\repeatthanks}


\authorrunning{M. Deutges et al.}

\institute{Institute of AI for Health, Helmholtz Zentrum München – German Research Center for Environmental Health, Neuherberg, Germany \and 
TUM School of Computation, Information and Technology, Technical University Munich, Munich, Germany \and 
Institute of Chemical Epigenetics, Faculty of Chemistry and Pharmacy,  Ludwig Maximilian University, Munich, Germany \and
Computer Aided Medical Procedures, Technical University of Munich, Munich, Germany}
    
\maketitle              
\begin{abstract}
The detection and segmentation of white blood cells in blood smear images is a key step in medical diagnostics, supporting various downstream tasks such as automated blood cell counting, morphological analysis, cell classification, and disease diagnosis and monitoring. Training robust and accurate models requires large amounts of labeled data, which is both time-consuming and expensive to acquire. In this work, we propose a novel approach for weakly supervised segmentation using neural cellular automata (NCA-WSS). By leveraging the feature maps generated by NCA during classification, we can extract segmentation masks without the need for retraining with segmentation labels. We evaluate our method on three white blood cell microscopy datasets and demonstrate that NCA-WSS significantly outperforms existing weakly supervised approaches. Our work illustrates the potential of NCA for both classification and segmentation in a weakly supervised framework, providing a scalable and efficient solution for medical image analysis.


\end{abstract}
\section{Introduction}
The segmentation of white blood cells (WBCs) in blood smear images plays a crucial role in medical diagnostics and treatment. Accurate segmentation of WBCs enable a range of downstream applications, including automated blood cell counting, morphological analysis, and disease diagnosis. These analyses are particularly relevant for hematologic disorders such as leukemia, where abnormalities in WBC morphology, size, and distribution serve as key diagnostic indicators \cite{shahzad2024blood}.

Manual segmentation remains the gold standard in clinical practice due to its accuracy, but it is labor-intensive, time-consuming, and subject to inter- and intra-observer variability. To address these challenges, automated segmentation methods have gained increasing attention as a means to improve efficiency, consistency, and scalability. However, traditional deep learning-based segmentation models rely on pixel-wise annotations, which are costly and time-consuming to obtain, limiting their applicability in real-world medical settings.

This has motivated the development of weakly supervised approaches that aim to infer segmentation masks from less expensive or already available annotations, such as image-level class labels \cite{lee2021railroad,luo2020learning,madni2025fl,su2021context,wang2020weakly,wang2020self,xu2022multi,zhang2020causal}. These approaches are particularly valuable in medical imaging, where acquiring fully annotated datasets is costly and time-consuming.

A common strategy for weakly supervised segmentation is to leverage class activation maps (CAMs) \cite{zhou2016learning}, which highlight the most discriminative regions of an image for a given class. Methods such as GradCAM \cite{selvaraju2017gradcam} and attention-based approaches \cite{li2018tell,wang2020self} have been widely used to localize objects within an image. However, these methods often struggle with capturing complete object structures, as they primarily focus on the most salient features and may fail to capture the full extent of a white blood cell.

An additional challenge for existing approaches in white blood cell segmentation is the poor generalization across different datasets and clinical settings. Variations in microscope scanners, staining protocols, and lighting conditions, can introduce significant domain shifts \cite{salehi2022unsupervised}. These shifts make it difficult for models trained on one dataset to perform well on another.

Recently, neural cellular automata (NCA) have emerged as a promising class of models for various tasks, demonstrating strong generalization capabilities by leveraging self-organizing behaviour through iterative update rules \cite{deutges2024neural,kalkhof2023med,mordvintsev2020growing}. This behavior enables NCA to learn robust, spatially structured representations that are less sensitive to domain shifts \cite{deutges2024neural}. Additionally, NCA models are highly parameter-efficient while maintaining strong performance, making them well-suited for applications with limited resources.

NCA provide a powerful mechanism for iterative feature extraction by modeling spatial interactions between pixels \cite{deutges2024neural,yang2025hierarchical}. Unlike conventional models, NCA dynamically refine local representations over multiple iterations, allowing the emergence of structured feature maps. This property enables the extraction of high-quality segmentation masks without the need for explicit supervision. 

Here, we propose a novel weakly supervised NCA approach for white blood cell segmentation. Our method extracts segmentation masks from NCA-generated feature maps using principal component analysis \cite{jolliffe2002principal} and is evaluated on two diverse datasets. To assess the robustness of our approach, we further examine its cross-domain performance across three datasets, demonstrating its generalizability in different data distributions.
To foster reproducible research, we publish our source code at \url{https://github.com/marrlab/NCA-WSS}.

\section{Methods}
We propose a weakly supervised segmentation approach where we first train an NCA-based classifier using image-level labels and then apply a feature-based segmentation strategy to localize white blood cells. In the following, we describe the details of our NCA architecture, classification framework, and segmentation mask extraction process.

\subsection{NCA Backbone}
NCA operate through iterative local updates, where each cell $c \in \mathbf{R}^n$ adjusts its state based on its (e.g. $3 \times 3$) neighborhood $N_c$. In image processing tasks, a cell corresponds to a pixel across all feature channels. The NCA update rule consists of two main functions:

\begin{itemize}
    \item The perception function $f_p$ aggregates local information using two convolutional filters.
    \item The update function $f_u$ processes the perception output through a two-layer fully connected network with ReLU activations.
\end{itemize}

The computed update is stochastically applied to the cell state at time $t$
\begin{equation}
    c^{t+1} = c^t + \delta f_u(f_p(N_c)),
\end{equation}
where $\delta$ is a binary variable acting as a regularization method that ensures only a random subset of cells are updated per iteration.

At time step $t$, the full state of all cells forms a feature representation $S_t \in \mathbf{R}^{H\times W\times n}$, which evolves over multiple iterations. Here, $H$, $W$, and $n$ are height, width and number of channels. The NCA update function can be expressed as
\begin{equation}
    \text{NCA}_\phi : \mathbf{R}^{H\times W\times n} \rightarrow \mathbf{R}^{H\times W\times n}
\end{equation}
where $\phi$ represents the learnable parameters of the perception and update functions.

\subsection{Classification Architecture}
Following NCA-based feature extraction, we aggregate the feature maps into a single vector for classification. Specifically, we apply average pooling across the spatial dimensions to obtain a compact representation
\begin{equation} 
\text{pool} : \mathbf{R}^{H\times W\times n} \rightarrow \mathbf{R}^n.
\end{equation}

The pooled feature vector is then passed through a fully connected network to predict class probabilities. By training the model end-to-end, the NCA learns to extract informative, structured features that are effective for classification.

\subsection{Segmentation Mask Extraction}

\begin{figure}[t]
\centering
\includegraphics[width=\textwidth,page=1,trim=2.5cm 18cm 1.5cm 3cm,clip]{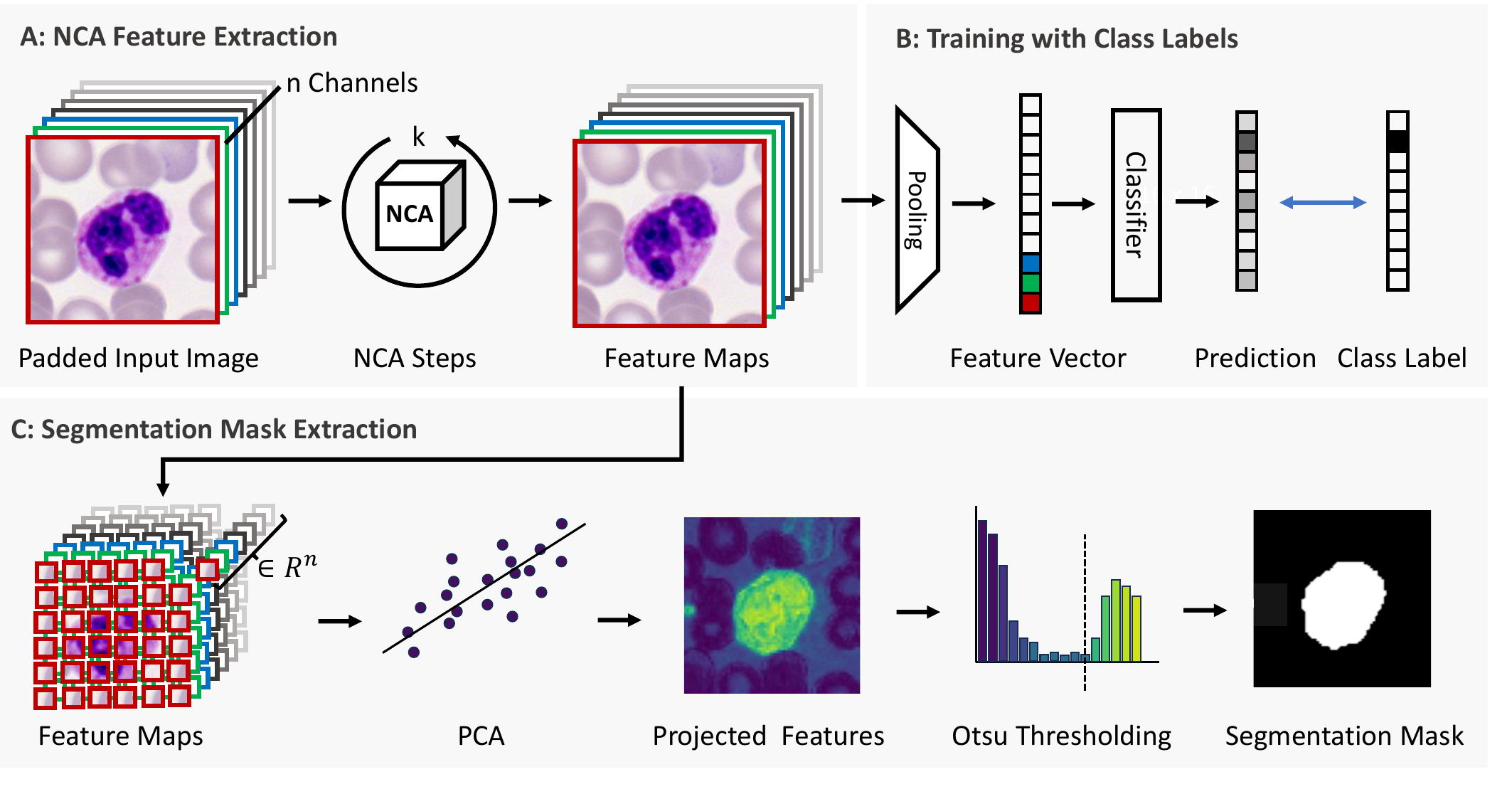}
\caption{Weakly Supervised Segmentation using Neural Cellular Automata (NCA-WSS) uses feature maps to identify WBCs in image patches. \textbf{A:} Feature maps are generated through the iterative local updates of the NCA. \textbf{B:} A class prediction is obtained by pooling NCA-generated feature maps, followed by a fully connected network. \textbf{C:} We extract segmentation masks by projecting the NCA features onto their first principal component, followed by Otsu thresholding to generate the binary mask.} 
\label{fig:fig1}
\end{figure}

Given an NCA-processed image with feature maps $S \in \mathbf{R}^{H \times W \times n}$, we treat each cell (i.e., each pixel across all $n$ feature channels) as a point in an $n$-dimensional space. We apply principal component analysis (PCA) \cite{jolliffe2002principal} to extract the principal direction of variance. Specifically, we compute the covariance matrix $\Sigma$ of the feature vectors across all spatial locations:
\begin{equation}
    \Sigma = \frac{1}{HW} \sum_{i,j} (S_{i,j} - \bar{S}) (S_{i,j} - \bar{S})^T, 
\end{equation}
where $\bar{S} \in \mathbf{R}^n$ is the mean of the NCA-cells. The first principal component $v_1 \in \mathbf{R}^n$ is obtained by solving
\begin{equation}
    v_1 = \arg\max_{v, \|v\|=1} v^T \Sigma v.
\end{equation}

We then project each feature vector onto $v_1$ to form a response map
\begin{equation}
    P_{i,j} = v_1^T S_{i,j}, \quad \forall (i,j) \in H \times W.
\end{equation}
This projection enhances high-variance regions, which correspond to informative structures within the WBC, while reducing background variations.

To obtain a binary segmentation mask, we apply Otsu’s thresholding \cite{otsu1975threshold} to $P$, selecting the optimal threshold $\tau^*$ that minimizes intra-class variance
\begin{equation}
    \tau^* = \arg\min_{\tau} \left( \omega_1(\tau) \sigma_1^2(\tau) + \omega_2(\tau) \sigma_2^2(\tau) \right),
\end{equation}
where $\omega_1, \omega_2$ and $\sigma_1^2, \sigma_2^2$ are the class probabilities and variances for pixels below and above the threshold $\tau$, respectively. The final segmentation mask is given by
\begin{equation}
    M_{i,j} = \mathbf{1}_{\{P_{i,j} > \tau^*\}}.
\end{equation}
This approach effectively combines the extracted feature representations learned by the classification model, which allows for segmentation without requiring annotations for segmentation.



\section{Experiments \& Results}
\subsection{Datasets}
We train our method using two datasets of white blood cell microscopy images, Raabin \cite{kouzehkanan2022large} and Matek19 \cite{matek2019human}. Images in both datasets are annotated with segmentation masks and class labels. Additionally, we collect an internal, independent dataset with segmentation masks but without class labels, serving as an extra test set to evaluate the generalizability of our weakly supervised segmentation method.

\begin{figure}[h]
\centering
\includegraphics[width=\textwidth,page=4,trim=0cm 25.5cm 0cm 0.5cm,clip]{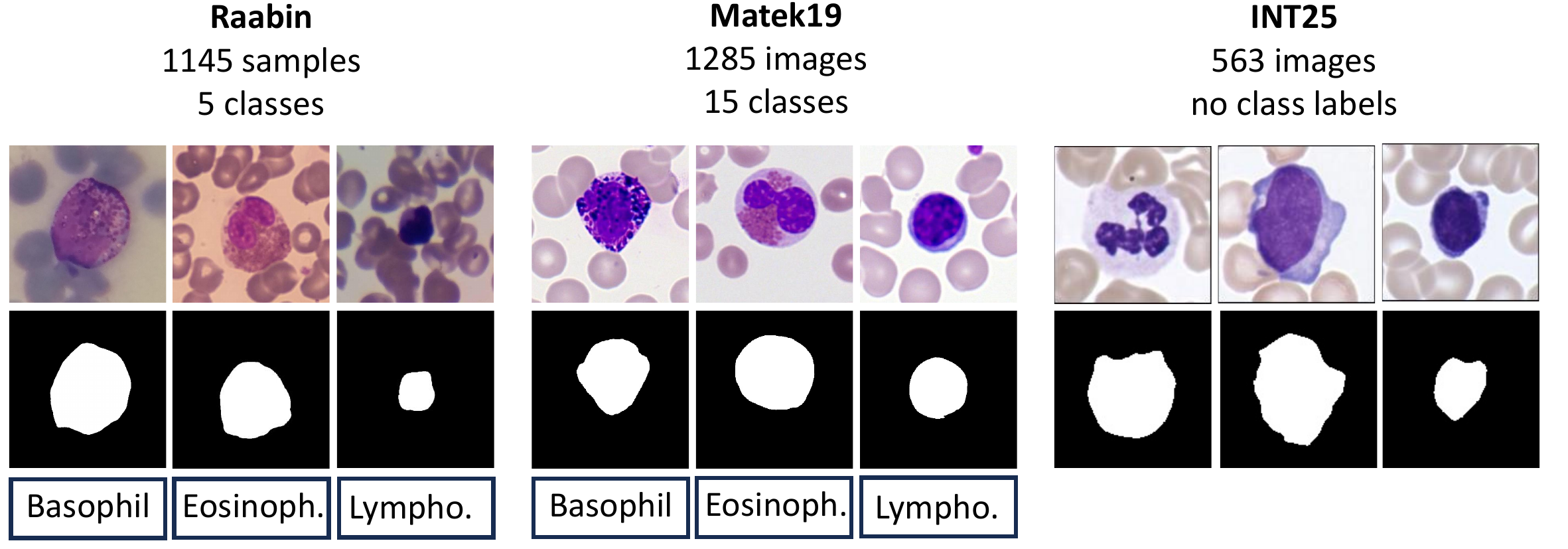}
\caption{Example images from the three white blood cell datasets used in this study. For each sample, we display the original blood smear image with its corresponding ground truth segmentation mask and class label below.} 
\label{fig:fig2}
\end{figure}

\paragraph{Raabin} \cite{kouzehkanan2022large} is a diverse dataset of white blood cell images covering five classes. The images are captured using different cameras and microscopes to ensure variability. For our experiments, we use a subset of 1,145 images that includes expert annotated segmentation masks.

\paragraph{Matek19} \cite{matek2019human} contains white blood cell images collected from 200 patients at the Munich University Hospital, with some images originating from patients diagnosed with acute myeloid leukemia. The dataset includes 15 distinct white blood cell classes, annotated by medical experts. For our experiments, we use a subset of 1,285 images that include both segmentation masks and class labels.

\paragraph{INT25} is our in-house dataset, consisting of 563 microscopy images including segmentation masks but no class labels. As a result, this dataset is used exclusively for evaluating segmentation performance in the cross domain experiment.

\subsection{Implementation Details}
\subsubsection{Model Parameters}
Our NCA backbone consists of 32 channels and a hidden layer of size 32. We iterate for 32 update steps before applying average pooling. The classifier network has a hidden layer of size 128.
\subsubsection{Training Parameters}
We optimize the model using an Adam optimizer with a learning rate of \( 10^{-4} \) and betas set to (0.9, 0.999). Learning rate decay is applied via an exponential scheduler with a decay rate of 0.9999. The model is trained using focal loss \cite{lin2020focalloss} with a batch size of 32. During training, we augment the images with random rotation and flips. We train for 256 epochs on the Raabin dataset and 128 epochs on Matek19. All experiments are conducted with five fold cross validation.

\subsection{Quantitative Results}
We evaluate the performance of our proposed method using the Intersection over Union (IoU) metric on the two datasets Matek19 and Raabin. For comparison, we include several baselines from a recently published study \cite{madni2025fl} where the authors trained the models in a federated learning setup using class labels. 
In contrast, our method trains on a single dataset and tests on the same set, without the benefit of federated learning's shared weights. 
This difference in training should not be expected to have an advantage due to the limited amount of data available for training compared to the federated learning setup that uses three datasets simultaneously. 

Table \ref{tab:iou} shows the IoU scores of the proposed NCA-WSS approach compared with the mentioned baselines when trained on both datasets.

\begin{table}[ht]
\caption{Our NCA-based weakly supervised segmentation significantly outperforms existing approaches in terms of IoU scores on both datasets. 
Mean and standard deviation are computed from five independent runs.}
\centering
\begin{tabular}{p{2.4cm}|P{2.3cm}|P{2.3cm}|P{2.3cm}}
Method & Backbone & IoU Raabin & IoU Matek19 \\ 
\hline
AuxSegNet \cite{xu2021leveraging} & ResNet38 & 37.2 ± 0.9 & 44.1 ± 0.1 \\
SEAM \cite{wang2020self} & ResNet38 & 33.6 ± 0.5 & 41.0 ± 0.7 \\
EPS \cite{lee2021railroad} & ResNet101 & 36.7 ± 0.4 & 45.7 ± 0.8 \\
Luo et al. \cite{luo2020learning} & VGG16 & 36.9 ± 0.4 & 41.9 ± 0.4 \\
CDA \cite{su2021context} & ResNet38 & 37.2 ± 0.2 & 43.5 ± 0.4 \\
MCTformer \cite{xu2022multi} & ResNet38 & 37.0 ± 0.8 & 44.8 ± 0.5 \\
Wang et al. \cite{wang2020weakly} & VGG16 & 31.9 ± 0.9 & 38.9 ± 0.2 \\
CONTA \cite{zhang2020causal} & ResNet38 & 33.8 ± 0.7 & 40.6 ± 0.5 \\
FL-W3S \cite{madni2025fl} & ResNet38 & 39.8 ± 0.5 & 47.0 ± 1.6 \\
\textbf{NCA-WSS} & \textbf{NCA} & \textbf{49.6 ± 1.8} & \textbf{82.6 ± 2.0} \\
\end{tabular}
\label{tab:iou}
\end{table}

To further assess generalization, we conduct a cross-domain experiment where we train on one dataset and test on the other two. Our method still achieves significantly superior performance compared to the baselines, even in this cross-domain setting. This result suggests that our model has good generalization capabilities, enabling it to perform well on unseen domains. Table \ref{tab:cross} shows the in-domain and cross-domain performances.

\begin{table}[ht]
\caption{Our NCA-based weakly supervised segmentation generalizes well to unseen domains. Mean and standard deviation of IoU scores are computed from five independent runs.}
\centering
\begin{tabular}{p{2cm}|P{2cm}|P{2cm}|P{2cm}}
\diagbox[width=2.1cm,height=0.6cm,innerleftsep=0.05cm,innerrightsep=0.05cm]{Train}{Test} & Raabin & Matek19 & INT25 \\ 
\hline
Raabin & 49.6 ± 1.8 & 64.9 ± 1.4 & 27.6 ± 6.1 \\
Matek19  & 55.1 ± 2.0 & 82.6 ± 2.0 & 63.7 ± 4.3 \\
\end{tabular}
\label{tab:cross}
\end{table}

\subsection{Discussion}

\begin{figure}[h]
\centering
\includegraphics[width=\textwidth,page=5,trim=1cm 13.5cm 1.7cm 1cm,clip]{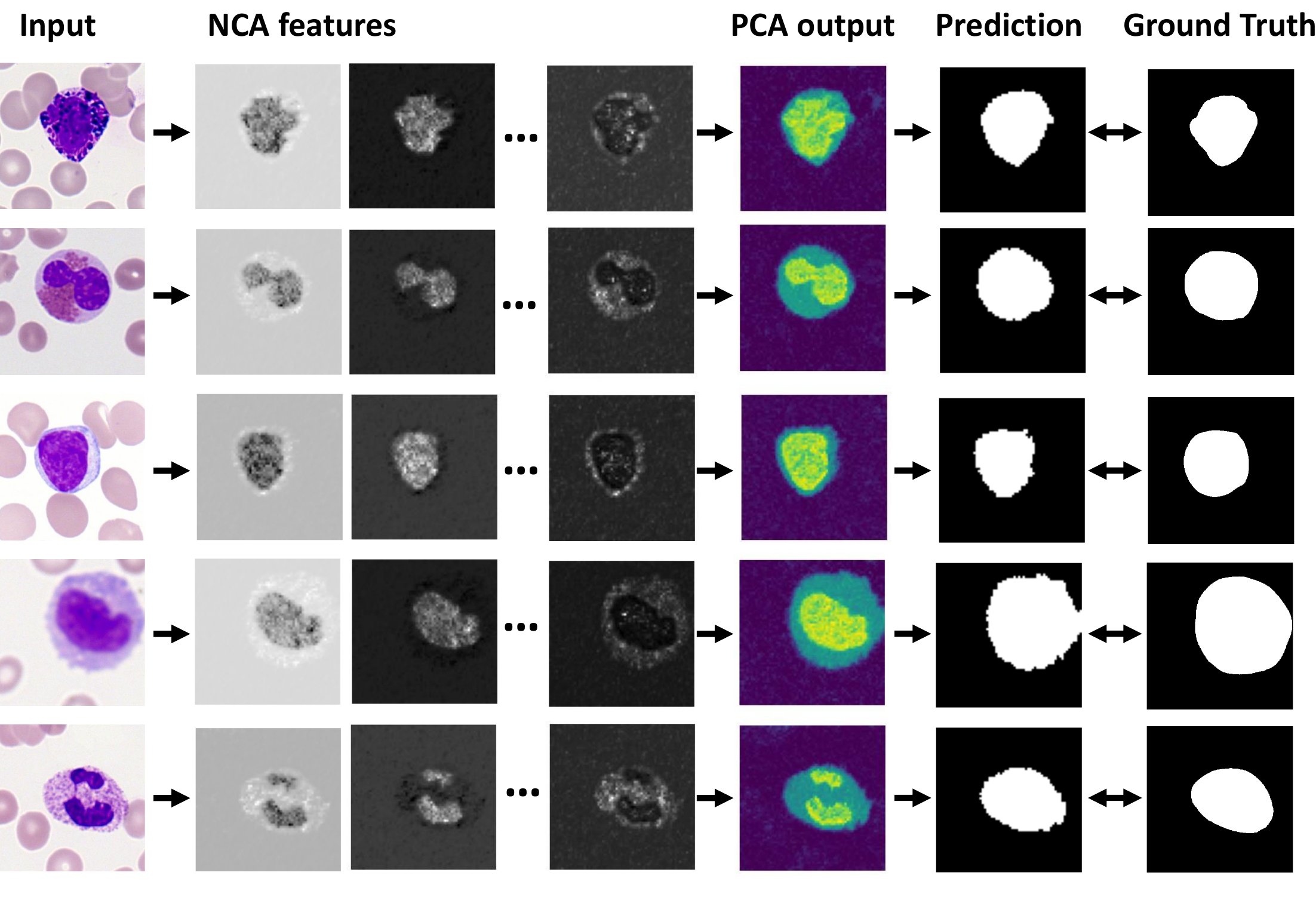}
\caption{Example of the extracted features and generated masks. The unique features learned by the NCA backbone allow for accurate segmentation of white blood cells. PCA highlights regions of high variance that correspond to the most relevant features within the cell, while filtering out noise.} 
\label{fig:fig3}
\end{figure}

In contrast to the existing approaches, our method fundamentally differs by leveraging the unique features extracted by the NCA backbone, which are then robustly combined using PCA (Figure \ref{fig:fig3}).

The NCA extracts localized features from the input image, focusing on discriminative information guided by the training with class labels. These features allow the model to effectively differentiate between cellular structures. Rather than directly relying on the absolute values of individual features, we use PCA to identify regions of high variance that typically correspond to the key structures within the white blood cell, such as the nucleus and cytoplasm.

By applying PCA, our model dampens the influence of background noise and non-informative variations, ensuring that only the most informative and stable features are used for segmentation. This focus on the most prominent features in terms of variance, rather than raw pixel values, contributes significantly to the higher IoU scores observed in our experiments, particularly in cross-domain scenarios.

These results highlight the robustness and effectiveness of our approach for weakly supervised semantic segmentation of WBCs, even when compared to state-of-the-art methods. Despite being trained on only a single dataset at a time, our method significantly outperforms all baselines trained using a federated learning scheme with three datasets, demonstrating its effectiveness even with limited training data.

\section{Conclusion}
In this work, we proposed a novel method for weakly-supervised segmentation of white blood cells, leveraging the unique capabilities of NCA combined with PCA. Our approach stands out by extracting highly localized and discriminative features through NCA, which are then robustly aggregated using PCA to obtain segmentation masks from a classification model without retraining.

Our experiments on three datasets demonstrate the effectiveness of our method in both in-domain and cross-domain scenarios, achieving superior segmentation results when compared to several baseline approaches. The robustness and high accuracy of our method highlight its potential for real-world applications in medical image analysis, where labeled segmentation data is scarce.

This work opens up new avenues for further improvements in weakly-supervised learning, and future efforts will explore extending the method to more complex segmentation tasks and broader datasets.

    

\begin{credits}
\subsubsection{\ackname} C.M. acknowledges funding from the European Research Council (ERC)
under the European Union's Horizon 2020 research and innovation program (Grant Agreement No. 866411 \& 101113551 \& 101213822) and support from the Hightech Agenda Bayern.

\subsubsection{\discintname}
The authors have no competing interests to declare that are relevant to the content of this article.
\end{credits}

%
%
%
\bibliographystyle{splncs04}
\bibliography{Paper-39}
\nocite{yang2025hierarchical}
%

\end{document}